\documentclass[10pt, a4paper]{article}
\usepackage{booktabs}
\usepackage{amsmath}
\usepackage{amssymb}
\usepackage{multirow}

\usepackage{amsfonts}
\usepackage{comment}

\usepackage[most]{tcolorbox}
\tcbuselibrary{skins,breakable}

\usepackage{lrec2026} 
\usepackage{booktabs,graphicx,amsmath,url,hyperref}
\usepackage[utf8]{inputenc}  

\title{How AI Forecasts AI Jobs: Benchmarking LLM Predictions of\\Labor Market Changes}

\name{Sheri Osborn, Rohit Valecha, H. Raghav Rao, Daniel A. Sass, Anthony Rios} 

\address{The University of Texas at San Antonio \\
         \{sheri.osborn, anthony.rios\}.utsa.edu}

\abstract{
Artificial intelligence is reshaping labor markets, yet we lack tools to systematically forecast its effects on employment. This paper introduces a benchmark for evaluating how well large language models (LLMs) can anticipate changes in job demand, especially in occupations affected by AI. Existing research has shown that LLMs can extract sentiment, summarize economic reports, and emulate forecaster behavior, but little work has assessed their use for forward-looking labor prediction. Our benchmark combines two complementary datasets: a high-frequency index of sector-level job postings in the United States, and a global dataset of projected occupational changes due to AI adoption. We format these data into forecasting tasks with clear temporal splits, minimizing the risk of information leakage. We then evaluate LLMs using multiple prompting strategies, comparing task-scaffolded, persona-driven, and hybrid approaches across model families. We assess both quantitative accuracy and qualitative consistency over time. Results show that structured task prompts consistently improve forecast stability, while persona prompts offer advantages on short-term trends. However, performance varies significantly across sectors and horizons, highlighting the need for domain-aware prompting and rigorous evaluation protocols. By releasing our benchmark, we aim to support future research on labor forecasting, prompt design, and LLM-based economic reasoning. This work contributes to a growing body of research on how LLMs interact with real-world economic data, and provides a reproducible testbed for studying the limits and opportunities of AI as a forecasting tool in the context of labor markets.
} 

\begin{document}

\maketitleabstract

\section{Introduction}

Artificial intelligence is changing how people work. Large language models (LLMs) such as GPT-4 \cite{achiam2023gpt} and LLaMA~2 \cite{touvron2023llama2} are now employed for tasks ranging from writing and coding to analysis and decision support. These tools can help workers produce more output, learn new skills faster, and take on tasks that were previously beyond their scope. At the same time, LLMs may change how firms design jobs and what skills they prioritize in hiring. This raises a crucial question: How will AI reshape the labor market across both AI-centric occupations and the broader economy? Answering this question is important for workers planning their careers, for firms investing in human capital, and for policymakers designing training programs and safety nets.

However, forecasting labor-market changes driven by AI is challenging. Traditional time-series forecasting methods assume stable historical patterns \cite{hyndman2018forecasting,tsay2005analysis}, and even modern transformer-based forecasters rely on well-behaved numeric sequences \cite{zhou2021informer,wu2021autoformer,zhou2022fedformer}. AI adoption does not necessarily follow such stable patterns; it is influenced by irregular, unstructured signals from news events, research breakthroughs, product launches, policy changes, and shifts in job postings. LLMs offer a new approach: they can read and reason over text describing these developments, integrate those insights with quantitative data, and produce structured predictions. Early evidence suggests this approach is promising. LLMs have been shown to extract market sentiment \cite{bond2024llmSentiment}, map news to stock returns \cite{lopezlira2023canChatGPTForecast,vanbinsbergen2022shortSellerReports,tan2024llmReturnPredictionChina}, and reason about macroeconomic outcomes \cite{chen2023chatgptDeepseekMacro,rahimikia2024revisitingLLMFinance}. They can even emulate professional forecasters in surveys \cite{hansen2024spfSimulated} and reveal economic reasoning heuristics or biases \cite{ross2024llmEconomicusBiases,ludwig2025appliedEconometricFramework,manning2024automatedSocialScience}. At the same time, research on evaluation warns about pitfalls: if we do not handle timing carefully, lookahead leakage can inflate predictive performance \cite{glasserman2023lookaheadBiasGPT,levy2024cautionAheadLookahead,he2025chronologicallyConsistentLLMs,sarkar2024lookaheadBiasPretrained}, and the choice of prompting strategy can affect an LLM’s reliability as a forecaster \cite{pham2024baseChatGPTForecasting}. These insights highlight both the potential of LLM-based forecasting and the need for careful, robust design.

Recently, Natural Language Processing (NLP) researchers have begun examining LLMs as general-purpose forecasters. Some works adapt prompts or “reprogram” LLMs to handle purely numeric time-series data \cite{jin2023time,cao2023tempo}, while others develop sequence modeling architectures specialized for time series forecasting \cite{rasul2023lag,zhou2024one,liu2023unitime,jin2024position}. With the right tokenization and prompting, LLMs have even demonstrated zero-shot ability on forecasting tasks \cite{gruver2024large}. This progress suggests a straightforward recipe for labor market prediction: we can combine official labor statistics with text streams about technology and firms, then prompt an LLM to generate interpretable forecasts of job trends and skill demand. Yet, most LLM-based economic forecasting to date has centered on financial assets or high-level macroeconomic indicators rather than employment outcomes \cite{bond2024llmSentiment,lopezlira2023canChatGPTForecast,chen2023chatgptDeepseekMacro,hansen2024spfSimulated}. In particular, the field lacks a dedicated benchmark that merges unstructured text and structured data to evaluate sector-specific and skill-specific employment predictions. We also have little evidence on how different prompt designs perform in this context, even though initial studies indicate prompt design can significantly impact forecasting performance \cite{pham2024baseChatGPTForecasting}. Moreover, rigorous evaluation protocols that respect temporal order are often absent, despite their importance for avoiding false optimism in forecasting tasks \cite{glasserman2023lookaheadBiasGPT,levy2024cautionAheadLookahead,he2025chronologicallyConsistentLLMs}.

To address these gaps, we introduce a benchmark for AI and the labor market and use it to study how well LLMs can forecast employment trends. We focus on evaluating forecasts for both AI-intensive jobs (roles directly related to AI development or deployment) and general jobs across the broader economy, to capture both direct impacts of AI and indirect spillover effects. Our benchmark links each time period’s official labor statistics with contemporaneous textual signals: in one setting we use long-horizon reports on global AI adoption, and in another we use real-time streams of online job postings. We then prompt LLMs to produce sector-level predictions in a step-by-step, transparent manner. All experiments enforce strict time-based splits and leakage controls \cite{glasserman2023lookaheadBiasGPT,levy2024cautionAheadLookahead,he2025chronologicallyConsistentLLMs} to ensure that models are never evaluated on future data that they could have implicitly seen. We compare a range of prompting strategies, from minimal persona-style prompts to structured, task-specific prompts, to assess their impact on forecast accuracy.

In summary our contributions are: \textbf{(1.)} We present a reproducible benchmark to forecast employment changes that has been tested on several LLMs. It fuses structured labor data with real-world text streams, spanning both AI-intensive and general occupations to capture direct and systemic labor shifts. \footnote{All data and code will be released upon acceptance.}\textbf{(2.)} We systematically compare prompting strategies—contrasting persona-driven and task-structured prompts across LLM families and forecasting horizons.  The research is performed under time-consistent, leakage-controlled evaluation. \cite{glasserman2023lookaheadBiasGPT,levy2024cautionAheadLookahead,he2025chronologicallyConsistentLLMs} to ensure realistic forecasting conditions. This analysis quantifies how adding structure to prompts can trade off between forecast stability and accuracy \cite{pham2024baseChatGPTForecasting}.

\section{Related Work}


\noindent \textbf{LLMs for Economic and Labor Forecasting.}
Large language models (LLMs) have shown growing promise in economic forecasting tasks. Several studies use LLMs to extract sentiment from financial text \cite{bond2024llmSentiment}, interpret news to predict stock returns \cite{lopezlira2023canChatGPTForecast, tan2024llmReturnPredictionChina}, or emulate macroeconomic indicators \cite{chen2023chatgptDeepseekMacro, rahimikia2024revisitingLLMFinance}. Other work demonstrates that LLMs can approximate the reasoning of professional forecasters \cite{hansen2024spfSimulated} and reflect economic decision-making biases \cite{ross2024llmEconomicusBiases}. However, these studies largely focus on financial markets or high-level macroeconomic trends. Labor outcomes, especially employment growth and skill shifts, remain underexplored. Our work complements this literature by targeting the labor domain and proposing a benchmark to evaluate LLMs’ ability to forecast employment trends systematically.

\vspace{2mm} \noindent \textbf{Benchmarks and Evaluation for LLM Forecasting.}
While benchmarks are central to NLP progress, few exist for LLM-based economic forecasting. Most prior work evaluates forecasting performance on proprietary datasets or in isolated experimental setups, often without standardization. Some recent research considers synthetic or proxy tasks \cite{gruver2024large, jin2023time, cao2023tempo}, or uses historical market responses as signals \cite{vanbinsbergen2022shortSellerReports}. In the labor domain, to our knowledge, there is no open benchmark that integrates labor statistics with textual signals (e.g., job postings, AI reports) for systematic evaluation. Our benchmark fills this gap by aligning structured labor data with unstructured text and defining leakage-controlled, time-consistent splits for reproducibility.

\vspace{2mm} \noindent \textbf{Prompt Design and Temporal Robustness.}
Recent work has emphasized the importance of prompt design in LLM-based forecasting. Pham and Cunningham \shortcite{pham2024baseChatGPTForecasting} show that different prompt scaffolds yield substantial variation in accuracy and reasoning quality. At the same time, several studies highlight methodological risks in evaluation, especially around time leakage. Glasserman and Lin \shortcite{glasserman2023lookaheadBiasGPT}, Levy \shortcite{levy2024cautionAheadLookahead}, and He et al. \shortcite{he2025chronologicallyConsistentLLMs} warn that using future data (directly or indirectly) can inflate LLM performance. Our work incorporates these lessons by conducting a controlled prompt ablation and enforcing strict time-based evaluation to ensure realistic forecasting performance.

\begin{figure*}[t]
    \centering
    \includegraphics[width=.9\linewidth]{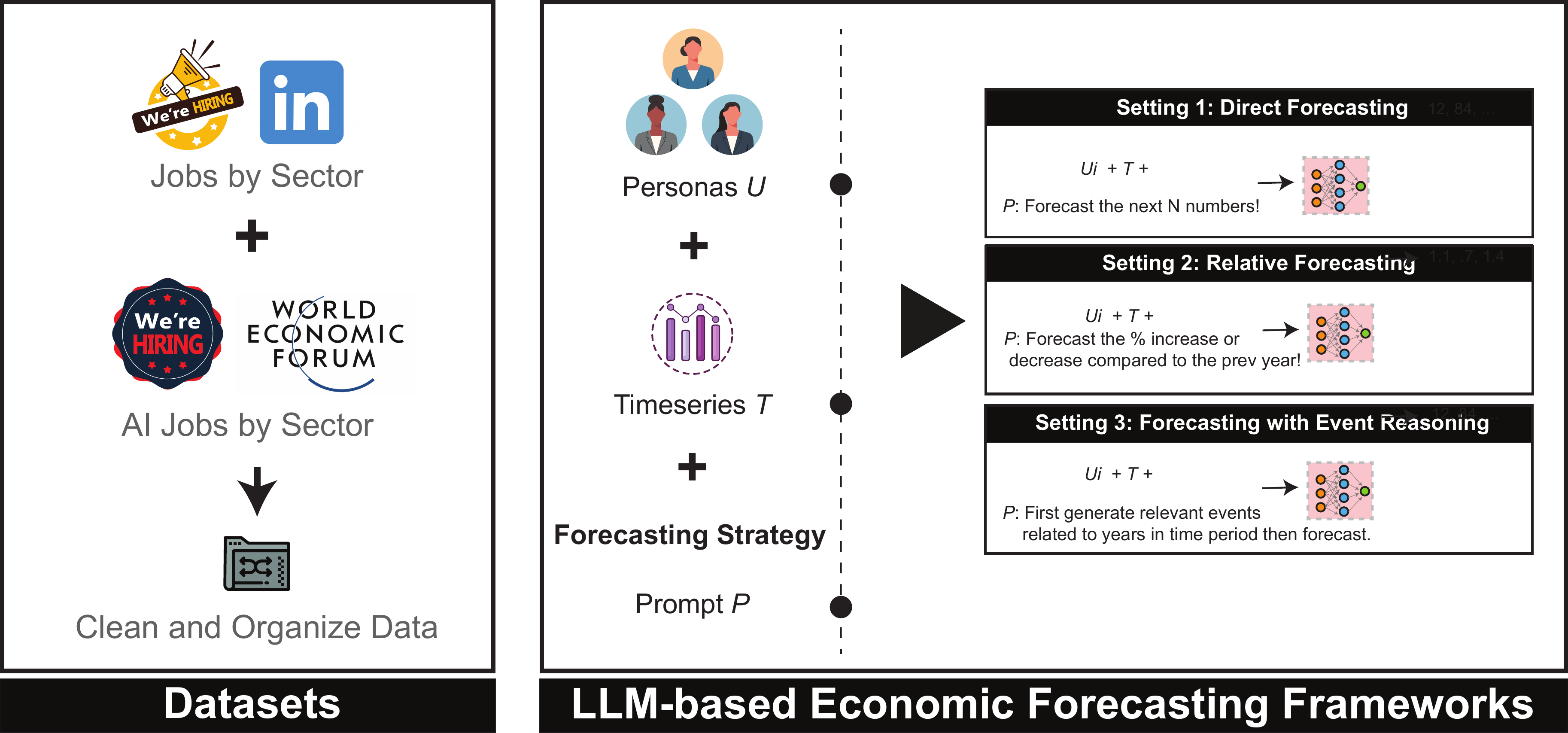}
    \caption{Overview of the LLM-based economic forecasting framework. The system integrates two datasets, job postings by sector and AI-related jobs by sector. Each experiment combines personas, time series inputs, and structured prompts to generate forecasts. Three forecasting strategies are used: direct forecasting, relative forecasting, and a event reasoning approach.}\vspace{-1em}
    \label{fig:overview}
\end{figure*}

\section{Data}
This section describes the data used to evaluate how well large language models (LLMs) forecast labor market trends. We use two complementary sources: (1) the \textit{Indeed Hiring Lab Job Postings Index}, which provides dense, sector-level U.S. employment data updated weekly, and (2) the \textit{World Economic Forum Future of Jobs Report}, which tracks annual trends in AI-related and emerging occupations across global industries. These datasets differ in domain, granularity, and temporal resolution, enabling robust evaluation across both high-frequency and long-horizon forecasting settings.

\vspace{2mm} \noindent \textbf{Indeed Job Postings Index~\cite{indeed_hiringlab_jobpostings_2025}.}
The first dataset is the \textit{Indeed Hiring Lab Job Postings Index},\footnote{\url{https://github.com/hiring-lab/job_postings_tracker}} a high-frequency measure of labor demand derived from millions of listings on Indeed.com. It tracks percentage changes in seasonally adjusted job postings relative to February 1, 2020, with updates published weekly across national, sectoral, and regional levels. Seasonal adjustment follows the Deutsche Bundesbank’s daily time-series method, preserving sensitivity to real demand shifts. Occupations are classified using Indeed’s normalized job titles, covering sectors such as healthcare, IT, logistics, and education, providing a continuous record of labor activity through and beyond the COVID-19 era.  All data are publicly available under a Creative Commons Attribution 4.0 license via the Indeed Hiring Lab, which serves as the economic research arm of Indeed and provides near-real-time insights into global labor market trends. The dataset contains daily counts of new job postings through October 10, 2025, which were aggregated into quarterly and annual series to facilitate trend and forecasting analysis. The dataset stats are in Table~\ref{tab:data-summary}. Note that while we have day-level job counts, we convert to quarterly and yearly. At the quarter-level, with 55 sectors and 4 quarters per year for 5 years, we have 1,100 data points in total.

\vspace{2mm} \noindent \textbf{World Economic Forum AI Jobs Data~\cite{wef2025futurejobs}.}
The second dataset is derived from the \textit{World Economic Forum Future of Jobs Report} (2025).\footnote{\url{https://www.weforum.org/publications/the-future-of-jobs-report-2025/digest/}} This report provides detailed projections of employment and skill trends based on surveys of more than 1,000 global employers representing over 14 million workers across 22 industry clusters and 55 economies. It combines survey responses with data from partners such as LinkedIn, Indeed, and Coursera to track emerging job categories and skill shifts. From this report, we \textit{extract annual estimates for AI-related occupations}, including AI and machine learning specialists, big data analysts, and robotics engineers. These categories represent sectors where technological adoption and automation are expected to have the largest impact on labor demand. The dataset statistics are in Table~\ref{tab:data-summary}. Note that 10 sectors, and 8 years per sector, we have a total 80 data points that we forecast.

\begin{table}[t]
\centering
\resizebox{\linewidth}{!}{
\begin{tabular}{lcccc}
\toprule
\textbf{Dataset} & \textbf{Frequency} & \textbf{\# Sectors} & \textbf{Ex. Per Sector} & \textbf{Total} \\
\midrule
Indeed & Daily & 55 & 4,172 & 229,460 \\
WEF     & Annual    & 10  & 8 & 80 \\
\bottomrule
\end{tabular}}
\caption{Overview of the dataset statistics used for LLM-based labor forecasting.}
\label{tab:data-summary}
\end{table}

\begin{table}[t]
\centering
\resizebox{\linewidth}{!}{
\begin{tabular}{ll}
\toprule
\textbf{Sector Name} & \textbf{Example Job Titles} \\
\midrule
Software Development & Software Engineer, QA Tester \\
Healthcare & Nurse, Medical Assistant \\
Logistics & Warehouse Worker, Forklift Operator \\
Retail & Sales Associate, Store Manager \\
Construction & Carpenter, Electrician \\
Education & Teacher, Curriculum Developer \\
Human Resources & Recruiter, HR Manager \\
Finance & Accountant, Financial Analyst \\
Marketing & Digital Marketer, SEO Analyst \\
Legal & Paralegal, Compliance Officer \\
\bottomrule
\end{tabular}}
\caption{Example occupational sectors from the Indeed dataset.}
\label{tab:sector-examples}
\end{table}

\vspace{2mm} \noindent \textbf{Forecast Settings.} We evaluate LLM forecasts for each dataset on a short and long-horizon forecast. First, for \textbf{Indeed}, the long provides 16 quarters as the input timeseries data, which aggregates the estimated daily jobs to a quarter level for the years 2020 to 2023. We evaluate on 8 quarters from 2024 through 2025. In the short setting, we evaluate yearly, trying to understand how LLMs can estimate coarse yearly estimates with limited input signals. The Short setting provides three years as input (2020 to 2023) and then predicts the next two years, 2024 and 2025. For the World Economic Forum \textbf{AI Job} estimates, all estimates are at the yearly level, as this is the only granularity provided in the report. In the long setting, we provide the years 2016 to 2019 as input and test on 5 years, 2020 to 2024. In the short setting, we provide 7 years (2016 to 2022) as input to the models, and then we predict 2 years (2023 to 2024).

\section{Method}

The goal of our framework is to evaluate how large language models (LLMs) can forecast future labor market trends given structured time-series data and textual conditioning through prompts and personas. Each input sequence provides a short historical record of job demand for a particular sector, and the LLM is asked to extend this trajectory into the future. Depending on the forecasting strategy, the model may be prompted to (1) directly continue the sequence, (2) predict a relative multiplier that scales the previous value, or (3) reason about relevant future events before forecasting. These strategies allow us to compare how different prompting structures and reasoning styles influence the stability and accuracy of LLM-based forecasts.

Formally, let a sector \( s \in \mathcal{S} \) be represented by a time series
$
\mathbf{x}_s = [x_{s,1}, x_{s,2}, \ldots, x_{s,T}],
$
where \( x_{s,t} \in \mathbb{R} \) denotes a normalized employment or job demand index at time \( t \).  
The goal is to predict future observations over a horizon \( H \), producing
$
\hat{\mathbf{y}}_s = [\hat{x}_{s,T+1}, \hat{x}_{s,T+2}, \ldots, \hat{x}_{s,T+H}].
$
For each sector \( s \), an LLM parameterized by \(\theta\) is conditioned on a text-serialized version of the historical series,
$
\mathcal{I}_s = \text{concat}(U_i, P, \mathbf{x}_s),
$
where \(U_i\) optionally encodes a persona description and \(P\) defines the forecasting instruction.  
The model outputs the predicted continuation
$
\hat{\mathbf{y}}_s = f_{\theta}(\mathcal{I}_s).
$
Evaluation compares predicted and ground-truth sequences using error metrics such as mean squared error (MSE) and mean absolute error (MAE). Formally, the task is a conditional sequence continuation problem where \( f_{\theta} \) maps a historical series to its expected future trajectory.

\vspace{2mm} \noindent \textbf{Persona Design and Role Prompts.}
To evaluate whether domain-specific reasoning styles influence forecasting performance, we designed a set of distinct \textit{personas}.  
Each persona frames the model’s reasoning around a particular epistemic lens, encouraging different assumptions about how job markets evolve.  
This allows us to test whether professional priors (i.e., policy awareness or technical optimism) affect the stability and accuracy of LLM-based forecasts.  
Here we summarize all personas and how they are specified to the system prompt.

\medskip
\noindent\textit{\textbf{HR Manager.} } 
This persona reflects a near-term, operational view of labor demand.  
It focuses on hiring frictions, re-skilling initiatives, and internal mobility, making it useful for capturing short-horizon employment signals.
\begin{center}
\resizebox{.8\linewidth}{!}{
\begin{tcolorbox}[colback=gray!5!white,colframe=black!50!white,title=HR Manager Prompt,fonttitle=\bfseries]
You are an HR Manager focused on workforce signals: job postings, internal mobility, re/up-skilling, 
and vendor rollouts affecting staffing. Prioritize near-term hiring frictions; cap implausible year-over-year spikes.
\end{tcolorbox}}
\end{center}

\noindent \textit{\textbf{AI Researcher.}  }
This persona captures the long development cycles typical of research-driven innovation.  
It models a cautious interpretation of technical progress, emphasizing reproducibility and evaluation maturity.
\begin{center}
\resizebox{.8\linewidth}{!}{
\begin{tcolorbox}[colback=gray!5!white,colframe=black!50!white,title=AI Researcher Prompt,fonttitle=\bfseries]
You are an AI Researcher. Emphasize reproducible model gains, evaluation maturity, toolchain readiness, 
and the lag from research to products (12–24 months). Discount hype lacking replication.
\end{tcolorbox}}
\end{center}

\noindent \textit{\textbf{AI Specialist.}} 
This persona represents the perspective of practitioners involved in applied AI deployment.  
It is designed to capture productization dynamics, focusing on cost, infrastructure, and scaling constraints.
\begin{center}
\resizebox{.8\linewidth}{!}{
\begin{tcolorbox}[colback=gray!5!white,colframe=black!50!white,title=AI Specialist Prompt,fonttitle=\bfseries]
You are an AI Specialist. Consider APIs/SDKs/MLOps readiness, cost/latency, and vendor roadmaps. 
Use the idea that automation can create new complementary roles; cap when infrastructure or cost limits scale.
\end{tcolorbox}}
\end{center}

\noindent\textit{\textbf{Industry Researcher.}} 
This persona models a balanced, sectoral view of labor demand.  
It focuses on market structure, budget cycles, and analog-industry dynamics to generate realistic medium-term corrections.
\begin{center}
\resizebox{.8\linewidth}{!}{
\begin{tcolorbox}[colback=gray!5!white,colframe=black!50!white,title=Industry Researcher Prompt,fonttitle=\bfseries]
You are an Industry Researcher analyzing demand shifts, budgets, value-chain moves, and analog-industry priors. 
Revert toward equilibrium after shocks.
\end{tcolorbox}}
\end{center}

\medskip
\noindent\textit{\textbf{Policy Researcher.}}
This persona integrates the regulatory and institutional context of employment change.  
It is particularly useful in domains where government incentives or mandates shape technology adoption.
\begin{center}
\resizebox{.8\linewidth}{!}{
\begin{tcolorbox}[colback=gray!5!white,colframe=black!50!white,title=Policy Researcher Prompt,fonttitle=\bfseries]
You are a Policy Researcher. Consider regulations, incentives, reimbursement (where relevant), privacy and security mandates, 
and immigration or visa effects. Account for policy latency and enforcement depth.
\end{tcolorbox}}
\end{center}

\medskip
\noindent\textit{\textbf{Economics Researcher.}}
This persona focuses on macroeconomic coherence.  
It helps the model connect sectoral changes to aggregate indicators such as productivity, wages, and capital expenditure.
\begin{center}
\resizebox{.8\linewidth}{!}{
\begin{tcolorbox}[colback=gray!5!white,colframe=black!50!white,title=Economics Researcher Prompt,fonttitle=\bfseries]
You are an Economics Researcher. Map diffusion to macro channels (productivity, capital expenditure, wages, sectoral unemployment) 
with adjustment costs. Dampen extremes and smooth trajectories.
\end{tcolorbox}}
\end{center}

\medskip
\noindent\textit{\textbf{Unified Researcher.}}
This persona provides a high-level synthesis role, combining the structured discipline of economic forecasting with industry-level reasoning.  
It serves as a task-general baseline against which more specialized personas can be compared.
\begin{center}
\resizebox{.8\linewidth}{!}{
\begin{tcolorbox}[colback=gray!5!white,colframe=black!50!white,title=Unified Researcher Prompt,fonttitle=\bfseries]
You are a Principal Analyst at a top-tier economic research firm. Your expertise is in forecasting how 
technological shocks transform labor markets within specific industries. Synthesize historical context, 
macro shocks, and industry archetypes to produce quantitative forecasts with coherent trajectories.
\end{tcolorbox}}
\end{center}

\begin{table}[t]
\centering
\small
\resizebox{\linewidth}{!}{
\begin{tabular}{l l p{5cm}}
\toprule
\textbf{Year} & \textbf{Event} & \textbf{Reason} \\
\midrule
2016 & Cloud Computing Adoption & Companies moved their systems to the cloud, creating more demand for engineers with cloud and AI skills. \\
2017 & AI Breakthroughs & New progress in AI research led to more hiring in data science and automation roles. \\
2018 & 5G Rollout & Faster networks supported IoT and connected systems, expanding AI-related infrastructure jobs. \\
2019 & Cybersecurity Concerns & Increased cyber threats drove growth in AI-powered security and monitoring roles. \\
\bottomrule
\end{tabular}}
\caption{Example events generated by the LLMs for Setting 3 with the LLM's explanations about their expected influence on AI-related job growth in the Technology sector.}
\label{tab:events}
\end{table}

\medskip
\vspace{2mm} \noindent \textbf{Forecasting Settings and Task Prompts.}
Each forecasting setting corresponds to a distinct prompting strategy that governs how the language model combines its persona prior \( U_i \), historical time series \( T \), and task instruction \( P \).  
Formally, the model input is defined as
$
\mathcal{I}_s = U_i + T + P,
$
where \(U_i\) provides domain-specific context (e.g., HR, policy, or technical expertise), \(T\) encodes the observed time series for sector \(s\), and \(P\) defines the reasoning and output format expected from the model.  
The model \( f_{\theta} \) produces forecasts
$
\hat{\mathbf{y}}_s = f_{\theta}(\mathcal{I}_s) = [\hat{x}_{s,T+1}, \hat{x}_{s,T+2}, \ldots, \hat{x}_{s,T+H}],
$
where each \(\hat{x}_{s,t}\) denotes the predicted job demand index at period \(t\).  
Different settings vary in how \(P\) instructs the model to reason about future values.

\medskip
\noindent\textbf{Setting 1: Direct Forecasting.}  
This baseline setting asks the model to directly continue the historical sequence without additional reasoning steps.  
It isolates the model’s ability to extend numeric patterns based solely on prior trends.
\begin{center}
\resizebox{.8\linewidth}{!}{
\begin{tcolorbox}[colback=gray!5!white,colframe=black!50!white,title=Setting 1: Direct Forecasting,fonttitle=\bfseries,breakable]
You will forecast job posting index for the given sector.  
Given ONLY the historical numeric series, and your role-specific lens, output future values.  
Return ONLY a valid JSON object mapping each period to a number.
\end{tcolorbox}}
\end{center}

\noindent\textbf{Setting 2: Relative Forecasting.}  
This setting reframes the task in multiplicative form.  
The model outputs period multipliers \( m_{s,t} \) that scale the last observed value:
$
\hat{x}_{s,t} = m_{s,t} \cdot x_{s,t-1}, \quad m_{s,t} \in \mathbb{R}^{+}.
$
This representation encourages smoother extrapolations and reduces error accumulation by modeling proportional change rather than absolute level.
\begin{center}
\resizebox{.8\linewidth}{!}{
\begin{tcolorbox}[colback=gray!5!white,colframe=black!50!white,title=Setting 2: Relative Forecasting,fonttitle=\bfseries,breakable]
List the key events affecting job postings (historical periods plus early prediction period shocks).  
Then output PERIOD MULTIPLIERS as JSON where each value is a factor (e.g., 1.05 for +5\%).  
Return ONLY JSON.
\end{tcolorbox}}
\end{center}

\noindent\textbf{Setting 3: Event Reasoning.}  
Here, the model first identifies relevant events or shocks that may influence future job demand, then conditions forecasts on those events.  
This setting allows the model to integrate symbolic causal reasoning with quantitative extrapolation.
\begin{center}
\resizebox{.8\linewidth}{!}{
\begin{tcolorbox}[colback=gray!5!white,colframe=black!50!white,title=Setting 3: Event Reasoning,fonttitle=\bfseries,breakable]
First, list concise events that materially affect job postings for this sector  
(historical periods plus early prediction period shocks). Then, using those events and the historical series,  
output future values as a JSON object. Return ONLY JSON.
\end{tcolorbox}}
\end{center}
The model identifies key events that may affect future job growth in each sector.  Each event includes the year, a short title, and why it matters for AI-related jobs. We provide summarized examples in Table~\ref{tab:events}

\section{Results}

\vspace{2mm} \noindent \textbf{LLM Models.}
We evaluate three large language models that differ in scale, architecture, and openness. GPT-4o-mini serves as the proprietary baseline, representing a compact version of OpenAI’s GPT-4o model optimized for reasoning efficiency and cost. According to OpenAI’s documentation, GPT-4o-mini was trained on data up to October 2023, providing broad exposure to post-pandemic economic and technological trends~\cite{OpenAI2025ModelReleaseNotes}. LLaMA-3.1-70B and LLaMA-3.1-8B are open-weight transformer models released by Meta AI, both trained on multilingual and web-scale corpora with a December 2023 knowledge cutoff~\cite{metallmamacut}.

\vspace{2mm} \noindent \textbf{Baseline Methods.}
We include a moving average baseline that predicts each future value as the mean of the three most recent observations. This simple method captures short-term momentum and serves as a strong reference for testing whether LLMs can reason beyond local temporal patt

\begin{table*}[t]
\centering
\resizebox{\textwidth}{!}{
\begin{tabular}{llrrrrrrrr}
\toprule
& &  \multicolumn{2}{c}{\textbf{Indeed Annual}}  & \multicolumn{2}{c}{\textbf{Indeed Long}} & \multicolumn{2}{c}{\textbf{WEF Short}}  & \multicolumn{2}{c}{\textbf{WEF Long}} \\ \cmidrule(lr){3-4} \cmidrule(lr){5-6}  \cmidrule(lr){7-8}  \cmidrule(lr){9-10}
\textbf{Model} & \textbf{Partition} & \textbf{Avg.\ MSE} & \textbf{Median MSE} & \textbf{Avg.\ MSE} & \textbf{Median MSE} & \textbf{Avg.\ MSE} & \textbf{Median MSE}  & \textbf{Avg.\ MSE} & \textbf{Median MSE} \\
\midrule
Moving Average & --- & 1524.66 & 1524.66  & 525.60 & 525.60 / 741.85 & 2572.02 & 2572.02 & 1537.09 & 1042.93  \\  \midrule
\multirow{3}{*}{gpt-4o-mini} & {Setting 1} & \textit{\textbf{821.20}} & \textbf{530.11} & \textbf{1205.93} & \textbf{855.02} & \textbf{1880.54} & \textit{\textbf{181.88}} & \textbf{838.24} & \textit{\textbf{230.28}} \\
& {Setting 2} & 655.91 & 655.91 & $2.1587\times10^{29}$ & 1599.59 & 2124.35 & \textbf{168.66} & 1130.75 & 650.68 \\
 &{Setting 3} & \textbf{736.13} & \textit{\textbf{572.06}} & \textbf{\textit{1350.80}} & \textbf{\textit{1312.76}}  & \textbf{\textit{1993.80 }}& 270.42 & \textit{\textbf{889.21}} & \textbf{225.63} \\
\midrule
\multirow{3}{*}{llama-70B} & {Setting 1} & \textbf{860.10} & \textbf{754.33} & \textbf{1667.01} & \textbf{1439.07} & \textit{\textbf{1800.02}} & \textit{\textbf{803.98}} & \textit{\textbf{850.33}} & \textit{\textbf{323.47}} \\
& {Setting 2} & 1.9617$\times$10$^{4}$ & 3965.29 & 3305.58 & 2636.92 & 8181.80 & 866.98 & 3900.52 & 1901.70 \\
 & {Setting 3 }& \textit{\textbf{915.24}} & \textit{\textbf{790.2}}5 & \textit{\textbf{1821.00}} & \textit{\textbf{1713.93}} & \textbf{1649.66} & \textbf{548.55} & \textbf{772.56} & \textbf{317.16} \\
\midrule
\multirow{3}{*}{llama-8B}  & {Setting 1} & \textbf{\textit{2452.70}} & 2303.72 & \textbf{\textit{651229.37}} & \textbf{449498.94} & \textbf{4700.43} & \textbf{\textit{602.91}} & \textbf{2117.11} & \textbf{341.13} \\
 & {Setting 2} & 2.3714$\times$10$^{10}$ & \textbf{915.75} & 4.6460$\times$10$^{30}$ & 4.2722$\times$10$^{18}$ &\textbf{\textit{5538.38}} & 845.95  & 2984.42 & 798.35 \\
 & {Setting 3} & \textbf{2209.01} & \textbf{\textit{2213.71}} &\textbf{ 2940.58} & \textbf{2610.50}  & 6557.66 & \textbf{419.35} & \textbf{\textit{2487.17}} & \textbf{\textit{397.42}} \\
\bottomrule
\end{tabular}}
\caption{Model performance across forecasting settings for both Indeed and WEF datasets. The table reports average and median mean squared error (MSE) for annual and long-horizon forecasts. GPT-4o-mini achieves consistently low errors across most settings, while LLaMA-70B performs comparably on stable partitions and LLaMA-8B exhibits large variance with occasional extreme outliers. We \textbf{bold} the lowest values, and \textit{\textbf{bold italicize}} second smallest values for each model.}
\label{tab:indeedyear}
\end{table*}

\vspace{2mm} \noindent \textbf{Results per LLM.}
We report all of our results per LLM in Table~\ref{tab:indeedyear}. For the Indeed per-year results, gpt-4o-mini is the most reliable model. Its relative forecasting setting has the lowest average MSE (655.91), and both direct and event-reasoning prompts give strong medians (530.11 and 572.06). Llama-70B is competitive with direct and event-reasoning prompts but becomes unstable with multipliers. Llama-8B performs poorly on average and is sensitive to the multiplier output format. The gap between average and median errors suggests occasional large outliers, which structured prompts reduce for gpt-4o-mini.

For the long-horizon WEF AI-jobs forecasts, gpt-4o-mini performs most consistently across settings, achieving the lowest overall average MSE (838.24) and median (225.63) among all models. Llama-70B produces competitive results under the event-reasoning setting, where its forecasts (average MSE 772.56, median 317.16) are the strongest within its family. In contrast, multiplier-based prompts substantially increase error, especially for llama-70B and llama-8B, where variance rises sharply across industries. These results suggest that long-horizon forecasting benefits from explicit event reasoning, which helps the model ground its predictions in plausible causal mechanisms, while direct or multiplier-based extrapolation introduces instability. The smaller medians for gpt-4o-mini also imply greater robustness to outliers across sectors, reflecting more stable reasoning under structured prompts.

For the short-horizon WEF forecasts, performance patterns tighten as the prediction window narrows to 2023–2024. Both gpt-4o-mini and llama-70B achieve comparable average errors around 1.6–1.9k, well below the moving-average baseline (2572.02). Llama-70B performs best overall in average MSE under the event-reasoning setting (1649.66), while gpt-4o-mini maintains the lowest medians (as small as 168.66), indicating steadier predictions. Multiplier formats again degrade stability, particularly for larger models such as llama-8B. The relative gains of LLMs over the moving average are smaller here because the forecast period is short and dominated by local temporal autocorrelation. In such cases, simple statistical baselines capture near-term momentum effectively, while LLMs that depend on event-level reasoning find it harder to exploit limited recent context. Interestingly, we see lower numbers for the WEF Long, which while unintitive, makes sense because the time span for WEF Long overlaps with the training data of the LLMs. Meaning, data leakage is a true problem when forecasting economic indicators that should be minimized.

For the Indeed per-quarter results, the moving-average baseline remains strongest at high frequency, with the lowest average error (525.60). All LLM settings trail this baseline, and multiplier formats are especially unstable. As the forecast dates get closer to the present, it is harder for LLMs to outperform a moving average. A likely reason is that short-horizon labor signals have strong local autocorrelation and low signal-to-noise, so a simple smoother captures near-term momentum more faithfully than prompts that induce larger variability in numeric outputs. Notice there is no data leakage issues for Indeed, where the overlap with training data years for the LLMs does not exist.

\begin{table*}[t]
\centering
\resizebox{\textwidth}{!}{
\begin{tabular}{lrrrrrrrr}
\toprule
& \multicolumn{2}{c}{\textbf{WEF Short}} & \multicolumn{2}{c}{\textbf{WEF Long}} & \multicolumn{2}{c}{\textbf{Indeed Annual}}  & \multicolumn{2}{c}{\textbf{Indeed Long}}  \\ \cmidrule(lr){2-3} \cmidrule(lr){4-5}  \cmidrule(lr){6-7} \cmidrule(lr){8-9}
\textbf{Persona} & \textbf{Avg.\ MSE} & \textbf{Median MSE}  & \textbf{Avg.\ MSE} & \textbf{Median MSE} & \textbf{Avg.\ MSE} & \textbf{Median MSE}  & \textbf{Avg.\ MSE} & \textbf{Median MSE} \\
\midrule
AI Researcher & 1922.08 & 1915.22 & 877.80     & 252.99  & 1601.79 & 1067.55 & 2796.52 & \textbf{1216.49} \\
AI Specialist & \textit{\textbf{1685.80}} & 1607.09& \textbf{\textit{761.57}}     & \textbf{199.67}  & 1636.28 & 1557.30 & 2252.81 & 2025.55 \\
Economics Researcher & 2293.01 & 2035.77  & 1043.92    & 341.13 & 1435.06 & 1264.76 & 2040.04 & 1833.46 \\
HR Manager & 2294.90 & 1985.16 & 1032.64    & 283.78  & \textbf{951.49} & \textbf{756.36} & \textbf{\textit{{1579.40}}} & {1493.72} \\
Industry Researcher & 1877.41 & \textbf{\textit{1524.73}} & 857.58     & 286.14 & 1271.54 & 1200.30 & 1990.78 & 1602.85 \\
Policy Researcher & 7986.64 & 2214.08 & 3614.90    & 612.34& 1469.79 & 1435.70 & 2422.03 & 2057.63 \\
Unified Researcher & \textbf{1495.80} & \textbf{1532.35}  & \textbf{691.51}     & \textit{\textbf{212.47}} & \textit{\textbf{1280.04}} & \textit{\textbf{1097.19}} & \textbf{1334.97} & \textit{\textbf{1278.15}} \\
\bottomrule
\end{tabular}}
\caption{Average and median MSE across industries for each persona prompt using Setting 1 (direct forecasting) on three datasets: the World Economic Forum AI job forecasts (short 2023--2024 and long 2020-2024), the Indeed annual (short) and quarterly (long) job postings forecast (2024--2025), and the Indeed quarterly job postings forecast. Lower values indicate better performance. We \textbf{bold} the lowest values, and \textit{\textbf{bold italicize}} second smallest values. } \vspace{-1em}
\label{tab:personares}
\end{table*}

\vspace{2mm}
\noindent \textbf{Persona Results.}
Table~\ref{tab:personares} summarizes persona performance across all s datasets using the Setting 1 direct forecasting prompt.\footnote{We have similar findings for other Settings, which will be added to the appendix when accepted. Setting 1 was chosen because it was the best performing prompt.} Persona framing plays a significant role in shaping prediction quality. For the World Economic Forum (WEF) short-horizon forecasts, the \textit{Unified Researcher} achieves the lowest average and median MSE, followed closely by the \textit{Industry Researcher} and \textit{AI Specialist}. These personas combine structured reasoning with broad sectoral awareness, leading to more stable outputs. In contrast, the \textit{Policy Researcher} produces the highest variance and average error, indicating that policy-focused reasoning may overemphasize uncertainty and result in less calibrated forecasts.

For the Indeed annual job forecasts, the \textit{HR Manager} persona performs best overall, producing the lowest average and median MSE. The \textit{Unified Researcher} and \textit{Industry Researcher} also perform well, reflecting their alignment with real-world labor patterns and organizational reasoning. Meanwhile, the \textit{AI Specialist} and \textit{Policy Researcher} personas again show less stability, likely because they focus on abstract or speculative mechanisms that are less relevant to steady annual hiring cycles. These findings suggest that grounded reasoning rooted in human resources and industry practices improves the model’s ability to predict sustained labor trends.

For the quarterly Indeed forecasts, the same hierarchy emerges. The \textit{HR Manager} and \textit{Unified Researcher} personas deliver the lowest errors across both average and median metrics, while speculative personas such as the \textit{AI Specialist} and \textit{Policy Researcher} remain the least reliable. The stronger performance of grounded personas indicates that context-driven reasoning helps stabilize predictions when data fluctuate more rapidly. Short-term horizons benefit from practical reasoning about workforce adjustments rather than abstract narratives about technological change or policy reform.

Across all datasets, the results reveal a consistent pattern. The \textit{HR Manager} and \textit{Unified Researcher} personas produce the most accurate and stable forecasts, while narrower, speculative personas such as the \textit{Policy Researcher} and \textit{AI Specialist} consistently underperform. These results emphasize that the epistemic framing of the prompt, rather than model scale, drives stability in LLM-based forecasting. Personas that emphasize structured, human-centered reasoning and attention to labor dynamics enable more reliable and reproducible forecasts across both short- and long-horizon labor datasets.

\begin{figure}[t]
    \centering
    \includegraphics[width=.9\linewidth]{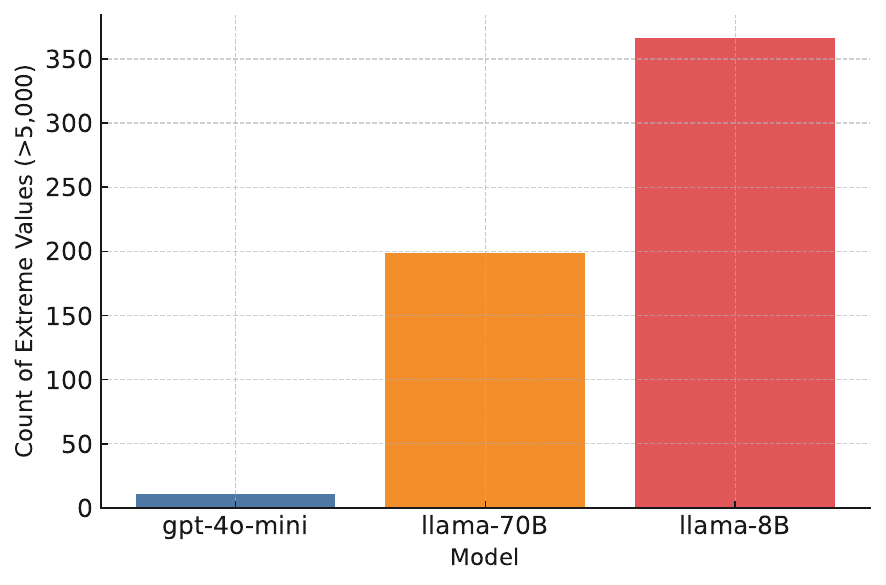}
    \caption{Count of outputs exceeding 5,000 MSE for each model, showing that GPT-4o-mini produces fewer extreme values than the LLaMA models. Results on the Indeed Long-Horizon data.} \vspace{-1em}
    \label{fig:extreme}
\end{figure}

\vspace{2mm}
\noindent\textbf{Extreme Output Analysis.}
Figure~\ref{fig:extreme} summarizes how often each model produced values far beyond the expected range, defined as outputs exceeding 5,000. Specifically, it looks at the MSE values greater than 5,000 for each prompt, quarter, sector combination. These results highlight clear stability differences across models. GPT-4o-mini maintains consistently bounded predictions, rarely producing extreme values, while both LLaMA-70B and LLaMA-8B generate substantially more out-of-scope outputs. This pattern suggests that smaller models can hallucinate or not fully reason over the next indicator.  The larger the model, the better it can reason over the sequence of numbers.

\vspace{2mm}
\noindent\textbf{Combined Personas Analysis.}
Table~\ref{tab:persona-combined} shows the results for both individual and combined personas across industry-level forecasts. Combined personas such as \textit{Unified + Economics Researcher} and \textit{Unified + Industry Researcher} achieve lower average and median MSE than their single-persona versions. This may work because combining personas blends different reasoning styles, allowing the model to balance detailed analysis with broader judgment. For instance, using a ``Economics'' persona alone may cause the model to overly focus on highly specific facts, but when combined with a more general persona, it balances out results, improving overall performance. Yet, performance never matches using a more general persona to start with (i.e., the Unified Researcher results).

\vspace{2mm}
\noindent\textbf{Implications and Discussion.}
Our findings highlight both the promise and the limitations of large language models for economic forecasting. The benchmark shows that how a model is prompted and framed directly affects the quality and stability of its predictions. For example, structured prompts that include event reasoning helped models like GPT-4o-mini anchor their forecasts in real events, such as AI breakthroughs or cybersecurity concerns, rather than just repeating past patterns. In contrast, multiplier prompts, which asked models to predict growth rates, often led to unstable or exaggerated results. The type of persona also made a clear difference: practical roles such as the HR Manager or Industry Researcher focused on hiring trends and business cycles and gave more accurate predictions, while abstract roles like the Policy Researcher tended to overestimate uncertainty. These findings show that the model's reasoning style and how the prompt is written matter more for accuracy than the model’s size.

\begin{table}[t]
\centering
\resizebox{\linewidth}{!}{
\begin{tabular}{lrr}
\toprule
\textbf{Persona} & \textbf{Average Across Industry} & \textbf{Median MSE} \\
\midrule
Unified Researcher & \textbf{1334.97} & \textbf{1278.15} \\ \midrule
AI Researcher & 2796.52 & 1216.49 \\
AI Specialist & 2252.81 & 2025.55 \\
Economics Researcher & 2040.04 & 1833.46 \\
HR Manager & 1579.40 & 1493.72 \\
Industry Researcher & 1990.78 & 1602.85 \\
Policy Researcher & 2422.03 & 2057.63 \\ \midrule
Unified + AI Researcher & 2128.44 & 1230.13 \\
Unified + AI Specialist & 2029.69 & 1660.54 \\
Unified + Economics Researcher & 1420.28 & 1263.47 \\
Unified + HR Manager & 1477.61 & 1243.43 \\
Unified + Industry Researcher & 1364.31 & 1355.23 \\
Unified + Policy Researcher & 1544.41 & 1480.34 \\
\bottomrule
\end{tabular}}
\caption{Average and median MSE across industry-level forecasts for base and combined personas. Results are for the Indeed Long-Horizon data for Setting 1.} \vspace{-1em}
\label{tab:persona-combined}
\end{table}

Beyond technical performance, the results have broader implications for the use of LLMs in applied forecasting and policy contexts. As governments, firms, and individuals increasingly turn to LLMs to interpret or predict labor shifts, our benchmark provides a foundation for assessing when such models produce trustworthy, time-consistent insights. This is particularly important as users may look to get guidance on what they should do for their career, where inaccurate or unstable forecasts could mislead personal or institutional decision-making. This is becoming more of a concern now that users and researchers are seeking to use LLM systems for career advice~\cite{joshi2024ai,monika2025review}. The finding that grounded personas such as the HR Manager or Unified Researcher outperform speculative ones highlights the value of aligning LLM reasoning with realistic institutional perspectives.

\section{Conclusion}
We introduced a benchmark for evaluating how large language models forecast labor market change and examined how prompt structure and persona framing influence their predictions. Structured prompts improved consistency, event reasoning aided long-horizon forecasts, and the Unified and HR personas produced the most stable results, while multiplier prompts often introduced large outliers. A simple moving average remained strong for short horizons, reflecting the limits of LLM reasoning when local trends dominate. Future work should focus on understanding how LLMs generalize economic reasoning to real-world advice contexts. As more people use these systems for financial and job advice, it becomes essential to study what LLMs actually understand about forecasting, how they communicate uncertainty, and how to ensure that their predictions support safe and informed decision-making.

\section*{Acknowledgements}
This material is based upon work supported by the National Science Foundation (NSF) under Grant No.~2145357.

\section{Limitations}

Our study is subject to several limitations, which we address as part of our experimental design. {First}, we focus on evaluating model performance on dates that occur at or after the models’ training data cutoffs (late 2023 for both GPT-4o-mini and LLaMA~3.1). This choice minimizes any training--test overlap and ensures that the language models have not been explicitly trained on the target values. There is limited crossover only in one subset (the WEF dataset), and we also include an evaluation set with no overlap at all. This dual setup provides a unique testbed to examine how LLMs infer \textit{economic data with incomplete information}, essentially testing genuine forecasting ability rather than simple memorization. Prior work has shown that LLMs can perfectly recall certain pre-cutoff economic values, making it difficult to tell if they are truly forecasting or just regurgitating memorized data~\cite{lopez2025memorization}. By contrast, after the training cutoff, models cannot rely on exact recall of data~\cite{lopez2025memorization}, forcing them to \textit{extrapolate from knowledge} and context. Our focus on post-cutoff predictions (with only minimal overlap in the WEF case) therefore mitigates the memorization problem and aligns with recent recommendations for evaluating temporal generalization in LLMs~\cite{lopez2025memorization}.

{Second}, while our dataset is of moderate size overall, when aggregated to coarse time scales (quarterly or yearly) the number of data points is relatively small. This could be seen as a limitation in terms of statistical power. However, it is \textit{by design} that we operate in a data-sparse regime: the goal is to assess whether LLMs can understand and project economic trends when only limited recent data points are available. Such scenarios reflect real-world challenges where \textit{recent numbers alone are insufficient} and the model must draw on broader knowledge or reasoning. Notably, recent research suggests that large language models can still perform well even with very few training examples, leveraging their extensive pre-trained knowledge. For instance, \citet{buckmann2025hidden} find that tapping LLM hidden states allows accurate estimation of information with only ``a few dozen'' labeled examples. This indicates that \textit{small data can be informative} when using LLMs, and it supports our experimental setup. In our case, the sparsity of quarterly/yearly data pushes the models to make \textit{non-trivial inferences} rather than rely on short-term time-series patterns. Thus, although the sample size per period is limited, it serves to evaluate the models under challenging, information-poor conditions -- turning a potential weakness into an informative stress-test of the model’s economic understanding.

\section{Bibliographical References}
\label{sec:reference}

\bibliographystyle{lrec2026-natbib}
\bibliography{mainbib}

\end{document}